\documentclass[12pt,twoside,a4paper]{article}
\usepackage{amsmath,amssymb,latexsym,theorem,natbib,epsfig,color,subfigure}
\usepackage{multirow,graphicx,array,rotating,url}  
\usepackage{epstopdf,afterpage,wasysym}
\usepackage{verbatim}
\usepackage[normalem]{ulem}
\newfont{\msa}{msam10 scaled\magstep1}
\newfont{\ssmsa}{msam9}

\def\crps{\mathop{\hbox{\rm CRPS}}}
\def\crpss{\mathop{\hbox{\rm CRPSS}}}

\def\md{\mathrm{MD}}

\numberwithin{equation}{section}

\newcommand{\revi}[1]{{\color{red}#1}}

\title{A two-step machine learning approach to statistical post-processing of weather forecasts for power generation}

\author{{\sc \'Agnes Baran} and {\sc S\'andor Baran} \vspace*{0.5cm}\\
         Faculty of Informatics, University of Debrecen\\
         Kassai \'ut 26, H-4028 Debrecen, Hungary
         }

        \date{}

\begin{document}
\pagestyle{myheadings}

\maketitle

\begin{abstract}
By the end of 2021, the renewable energy share of the global electricity capacity reached $38.3\,\%$ and the new installations are dominated by wind and solar energy, showing global increases of $12.7\,\%$ and $18.5\,\%$, respectively.  However, both wind and photovoltaic energy sources are highly volatile making planning difficult for grid operators, so accurate forecasts of the corresponding weather variables are essential for reliable electricity predictions. The most advanced approach in weather prediction is the ensemble method, which opens the door for probabilistic forecasting; though ensemble forecast are often underdispersive and subject to systematic bias. Hence, they require some form of statistical post-processing, where parametric models provide full predictive distributions of the weather variables at hand. We propose a general two-step machine learning-based approach to calibrating ensemble weather forecasts, where in the first step improved point forecasts are generated, which are then together with various ensemble statistics serve as input features of the neural network estimating the parameters of the predictive distribution. In two case studies based of 100m wind speed and global horizontal irradiance forecasts of the operational ensemble prediction system of the Hungarian Meteorological Service, the predictive performance of this novel method is compared with the forecast skill of the raw ensemble and the state-of-the-art parametric approaches. Both case studies confirm that at least up to 48h statistical post-processing substantially improves the predictive performance of the raw ensemble for all considered forecast horizons. The investigated variants of the proposed two-step method outperform in skill their competitors and the suggested new approach is well applicable for different weather quantities and for a fair range of predictive distributions.

\bigskip
\noindent {\em Key words:\/} 1D convolutional neural network, ensemble calibration, ensemble model output statistics, multilayer perceptron, solar iradiance, wind speed 
\end{abstract}

\section{Introduction}
\label{sec1}

The transition from fossil fuels to renewable energy sources is an essential step towards achieving the climate goals and fighting the challenges caused by emission of greenhouse gases and air pollution. By the end of 2021, the renewable energy share of the global electricity capacity reached $38.3\,\%$, whereas in the European Union this portion went up to $79.6\,\%$. The largest part of $40\,\%$ of the global renewable capacity is still corresponds to traditional hydro power; however, the newly installed capacity is dominated by wind and solar energy showing global increases of $12.7\,\%$ and $18.5\,\%$, respectively \citep{irena}. In 2021 wind farms covered $15\,\%$ of the total electricity demand of the EU and United Kingdom \citep{we22}; moreover, according to the independent climate think-tank Ember, during the peak months June and July 2021, for the first time, solar panels generated a tenth of EU electricity\footnote{Ember: EU solar power hits new record peak this summer.  Available at: \url{https://ember-climate.org/insights/research/eu-solar-power-hits-new-record-peak-this-summer/} [Accessed on 14 July 2022]}. However, both wind and photovoltaic energy sources are highly volatile making planning difficult for grid operators. Hence, according to the scheduling requirements e.g. in Hungary, power plants must indicate in advance in quarter-hour increments how
much energy they will produce, which makes accurate short-range prediction of wind and solar power to be of utmost importance.

Despite the complex relationship between wind speed/solar irradiance and produced wind/photovoltaic (PV) energy, accurate forecasts of the corresponding weather variables are essential for reliable electricity predictions. Weather forecasts are typically produced with the help of numerical weather prediction (NWP) models describing the dynamical and physical behaviour of the atmosphere with the help of nonlinear partial differential equations. In order to address uncertainties in the numerical model itself or in the initial conditions, NWP models are usually run several times with varying initial conditions or model parametrizations resulting in a forecast ensemble \citep{btb15,b18a}. In the last thirty years the ensemble method became a widely used approach all over the world, as the derived probabilistic forecasts allowing the estimation of probability distributions of future weather variables provide an important tool to help forecast-based decision making \citep{ffh19}. Recently, ensemble weather forecasts also serve as inputs to probabilistic hydrological \citep{cp09} or renewable energy forecasts \citep{pm18}.

However, despite the weather centres continuously improve their operational ensemble prediction systems (EPSs), ensemble forecasts still keep suffering from systematic errors such as lack of calibration or bias \citep{bhtp05}, thus require some form of statistical post-processing \citep{b18b}. Various approaches to statistical post-processing of a wide variety of weather variables have been proposed over the recent years; for a detailed overview of the state-of-the-art techniques see e.g. \citet{w18} or \citet{vbde21}.

The present article builds on parametric post-processing methods providing full predictive distributions of the weather variables at hand. A popular choice is the non-homogeneous regression or ensemble model output statistics \citep[EMOS;][]{grwg05}, which specifies the predictive distribution by a single parametric law with parameters connected to the ensemble members via appropriate link functions. EMOS models for different weather quantities mainly differ in the parametric family describing the predictive distribution; however, the link functions connecting the distribution parameters to the ensemble forecast might also differ. This computationally efficient method shows excellent performance for a large variety of EPSs, forecast domains and weather variables \citep[see e.g.][Section 8.3.2]{w19}; moreover, the majority of the existing EMOS models are implemented in the {\tt ensembleMOS R} package \citep{emos}.

Here we focus on quantities important in wind and solar energy production, and investigate statistical post-processing of ensemble forecasts of wind speed
measured at hub height (100m) and global horizontal irradiance (GHI). The presented case studies are based on 11-member short term ensemble forecasts of these variables produced by the Applications of Research to Operations at Mesoscale EPS \citep[AROME-EPS;][]{rvsz20} of the Hungarian Meteorological Service (HMS). Wind speed calls for EMOS predictive distributions with a non-negative support and positive skew, such as the truncated normal \citep[TN;][]{tg10}, the log-normal \citep[LN;][]{bl15}, or the truncated generalized extreme value \citep[TGEV;][]{bszsz21}, whereas the discrete-continuous nature of solar irradiance, similar to precipitation accumulation, requires non-negative predictive distributions assigning positive mass to the event of zero irradiance. The simplest method is to left-censor a suitable distribution at zero, which approach is followed by \citet{sch14} and \citet{bn16} in precipitation modelling, and recently by \citet{sealb21} proposing censored logistic (CL0) and censored normal (CN0) EMOS models for calibrating solar irradiance ensemble forecasts.

To improve the flexibility of parametric post-processing methods, \citet{rl18} and \citet{gzshf21, gzshf22} apply machine learning-based techniques for estimation of the unknown parameters of the predictive distributions for temperature and precipitation accumulation, respectively. A modified version of this approach appears in \citet{bb21}, where a novel model using a TN predictive distribution for calibrating hub height short term AROME-EPS wind speed ensemble forecasts with parameters connected to the ensemble members via a multilayer perceptron neural network \citep[MLP; see e.g.][Chapter 6]{dlbook} significantly outperforms the state-of-the-art EMOS methods. In contrast to  \citet{rl18} and \citet{gzshf21, gzshf22}, the proposed approach relies on the same training data as the corresponding EMOS models, that is neither additional input variables, nor large training data sets are required.

In the present paper we first extend the technique of \citet{bb21} to CN0 and LN predictive distributions, where the censored Gaussian law is applied for post-processing of solar irradiance, whereas with the LN model wind speed ensemble forecasts are calibrated.
Further, we suggest a new general two-step method for estimating the unknown parameters of the TN, LN and CN0 predictive distributions. Using the same training data as before, with the help of auxiliary neural networks we first provide improved predictions of the investigated weather variables. The required distribution parameters are then obtained in the second step, where the input features of the corresponding neural networks are extended with the previously obtained corrected point forecasts. The predictive performance of the proposed one- and two-step machine learning-based TN, LN and CN0 models is compared to that of the corresponding EMOS approaches with matching predictive distributions and to the raw AROME-EPS 100m wind speed and GHI ensemble forecasts, respectively.

\section{Data}
\label{sec2}

As mentioned, in the case studies of Section \ref{sec4} we consider two different weather variables used in renewable energy production, namely 100m  wind speed and GHI. Both investigated data sets contain ensemble forecasts of the AROME-EPS system of the HMS together with the corresponding validating observations. Wind speed data set is an extension of forecast-observation pairs investigated in \citet{bb21}, whereas part of the solar irradiance data has already been studied in \citet{sealb21}.

The 11-member AROME-EPS, consisting of a control run and 10 ensemble members obtained from perturbed initial conditions, covers the Transcarpatian Basin with a horizontal resolution of 2.5 km. For both studied weather quantities ensemble forecasts initialized at 0000 UTC with a forecast horizon of 48h and a temporal resolution of 15 minutes at the nearest grid point to the observation sites are available for the period between 7 May 2020 and 31 July 2021.

Validating  observations of 100m wind speed ($m/s$) with a 15-minute temporal resolution are given for three wind farms in the north-western part of Hungary (\'Acs, J\'anossomorja and P\'apakov\'acsi) for the period 1 January 2020 to 5 July 2021. The equal temporal resolution of forecasts and observations results in 192 forecast cases per run. While the AROME-EPS forecast data set for these three locations is complete, this is far not the case with data provided by the wind farms, as around 3\,\% of the observations are missing.

GHI observations ($W/m^2$) are available from two different sources. For modelling purposes we consider high quality observations of the HMS with a temporal resolution of 10 minutes for seven representative locations in Hungary (Asz\'od, Budapest, Debrecen, Kecskem\'et, P\'ecs, Szeged, T\'api\'oszele) covering the time interval between 1 April 2020 and 30 June 2021. However, we also have observations for two solar farms (Monor and Nagyk\H or\"os) with a 15-minute temporal resolution for the period 1 May 2020 to 18 August 2021. In the case study of Section \ref{sec4}, data for these two locations, where around 2\,\% of the observations are missing, are merely used for model verification. Due to the difference in the time steps of the AROME-EPS forecasts and HMS observations, in our analysis we consider GHI data with a temporal resolution of 30 minutes resulting in a total of 96 forecast cases per submission.

\section{Post-processing methods and forecast evaluation}
\label{sec3}

In what follows,  let \ $f_1,f_2, \ldots , f_{11}$ \ denote the 11-member AROME-EPS forecast for a given location, time point and forecast horizon, where \ $f_1=f_{\text{CTRL}}$ \ is the control forecast, whereas \ $f_2,f_3, \ldots ,f_{11}$ \ correspond to ensemble members \  $f_{\text{ENS},1},f_{\text{ENS},2}, \ldots ,f_{\text{ENS},10}$  \ generated using random perturbations. These latter $10$ forecasts are statistically indistinguishable and should be treated as exchangeable. Further, denote by  \ $\overline f$ \ the ensemble mean and by \ $\overline f_{\text{ENS}}$ \ the mean of the $10$ exchangeable members, while  \ $S^2$ \ and \ $\md$ \ stand for the ensemble variance and ensemble mean absolute difference, respectively, defined as
\begin{equation*}
S^2:=\frac 1{10}\sum_{k=1}^{11}\big(f_k-\overline f\big)^2 \qquad \text{and} \qquad \md:=\frac 1{11^2}\sum_{k=1}^{11}\sum_{\ell=1}^{11}\big|f_k-f_{\ell}\big|.
  \end{equation*}

\subsection{Ensemble model output statistics}
\label{subs3.1}

The following short review focuses on EMOS models showing the best forecast skill in the preliminary studies with the AROME-EPS 100m wind speed \citep{bb21} and GHI \citep{sealb21} ensemble forecasts.

\subsubsection{EMOS models for wind speed}
\label{subs3.1.1}

The first considered method to calibrate AROME-EPS hub height wind speed ensemble forecasts is a TN EMOS model with predictive distribution
\begin{equation}
    \label{tn_emos}
    \mathcal{N}_0\big(a_0 + a^2_{\text{CTRL}}f_{\text{CTRL}}+a^2_{\text{ENS}}\overline f_{\text{ENS}}, b^2_0 + b^2_1\md\big),
  \end{equation}
where \ $\mathcal{N}_0\big(\mu,\sigma^2\big)$ \ denotes a TN distribution  with location \ $\mu$ \ and scale \ $\sigma > 0$, \ left-truncated at $0$, having probability density function (PDF)
\begin{equation*}
    g(x|\mu,\sigma) := 
    \frac1{\sigma}\varphi\big((x-\mu)/\sigma\big) / \Phi(\mu/\sigma), \qquad \text{if \ $x\geq 0$,}
  \end{equation*}
and \  $g(x|\mu,\sigma):=0$, \ otherwise, with \ $\varphi$ \ and \ $\Phi$ \ being the PDF and the cumulative distribution function (CDF) of a standard normal distribution, respectively. According to the optimum score principle of \citet{gr07}, model parameters \ $a_0,a_{\text{CTRL}},a_{\text{ENS}},b_0,b_1\in {\mathbb R}$ \ are estimated by optimizing the mean value of a proper
scoring rule (see Section \ref{subs3.2}) over the forecast cases in the training data consisting of past forecast-observation pairs.

As an alternative, we also consider the LN EMOS approach of \citet{bl15}, where the mean \ $m$ \ and variance \ $v$ \ of the LN predictive distribution are linked to the ensemble members via expressions
\begin{equation*}
  m=\alpha _0 + \alpha^2_{\text{CTRL}}f_{\text{CTRL}}+\alpha^2_{\text{ENS}}\overline f_{\text{ENS}} \qquad \text{and} \qquad  v=\beta^2_0 + \beta^2_1S^2.
\end{equation*}
Similar to the TN EMOS model \eqref{tn_emos}, to obtain 
parameters \ $\alpha_0,\alpha_{\text{CTRL}},\alpha_{\text{ENS}},\beta_0,\beta_1\in {\mathbb R}$, \ one has to perform an optimum score estimation based on some verification measure.

\subsubsection{EMOS models for solar irradiance}
\label{subs3.1.2}

Consider a logistic distribution with location \ $\mu$ \ and scale \ $\sigma>0$ \ specified by the CDF
\begin{equation*}
  G(x|\mu,\sigma):=\big(1+{\mathrm e}^{-(x-\mu)/\sigma}\big)^{-1}, \qquad x\in {\mathbb R}.
\end{equation*}
Then the CDF of the logistic distribution with parameters \ $\mu$, \ and  \ $\sigma$ \ left-censored at zero (CL0) is given by
\begin{equation*}
  G_0^c(x|\mu,\sigma):=\begin{cases} G(x|\mu,\sigma),& \quad x\geq 0, \\
    0, & \quad x<0. \end{cases}
\end{equation*}

In the CL0 EMOS model of \citet{sealb21}, the link functions connecting the location parameter \ $\mu$ \ and the scale parameter \ $\sigma$ \ to the ensemble members are given by
\begin{equation}
  \label{cl0_emos}
  \mu = \gamma _0 + \gamma_{\text{CTRL}}f_{\text{CTRL}} + \gamma_{\text{ENS}}\overline f_{\text{ENS}} + \nu p_0 \qquad \text{and} \qquad \sigma = \exp\big(\delta_0 + \delta_1 \log S^2\big),
\end{equation}
where \ $p_0$ \ is the proportion of zero forecasts in the ensemble, that is
\begin{equation*}
  p_0:=\frac 1{11}\sum_{k=1}^{11}{\mathbb I}_{\{f_k=0\}}
\end{equation*}
with \ ${\mathbb I}_H$ \ denoting the indicator function of a set \ $H$. \ As before, model parameters \ $\gamma_0,\gamma_{\text{CTRL}},\gamma_{\text{ENS}},\nu, \delta_0,\delta_1\in {\mathbb R}$ \ are estimated by optimizing an appropriate verification measure over the training data.

As an alternative predictive distribution, one can also consider a normal law with mean \ $\mu$ \ and standard deviation \ $\sigma$ \ left-censored at zero (CN0) and use again the link functions specified by \eqref{cl0_emos}. Note that this model, referred to as CN0 EMOS, have already been mentioned in \citet{sealb21} and in the case study of Section \ref{subs4.2} the CL0 and CN0 EMOS approaches exhibit almost equal predictive performance.

\subsection{Parameter estimation}
\label{subs3.2}

All calibration methods are based on training data of forecast-observation pairs, and the appropriate temporal and spatial composition of this data set is an important issue in  statistical post-processing. In the case studies of Section \ref{sec4} we consider rolling training periods, where parameter estimates are obtained using ensemble forecasts and corresponding validating observations from the preceding \ $\ell_{tp}$ \ calendar days. This temporal composition is the standard approach in distribution-based post-processing, and the optimal training period length is usually determined after some preliminary data analysis. For the EMOS models of Section \ref{subs3.1} we treat the different forecast horizons separately, which is in full accordance with the approaches of  \citet{bb21} and \citet{sealb21}. Once the temporal range of the training data is given, one can choose between two traditional approaches to spatial selection \citep{tg10}. In the regional (or global) approach, model parameters are estimated using training data of the whole ensemble domain and all locations share the same set of parameters. It requires quite short training periods and allows an extrapolation of the predictive distribution to unobserved locations where ensemble forecasts are available. In contrast, local parameter estimation results in distinct parameter estimates for the different stations obtained using only training data of the given station. To avoid numerical stability problems, local models require much longer training periods \citep[for optimal training period lengths for EMOS modelling of different weather quantities see e.g.][]{hspbh14}, but if the training data is large enough, local models will usually outperform their regional counterparts.

As mentioned, the optimum score estimates of the unknown parameters of the various EMOS models minimize the mean value of a scoring rule over
the training data. Scoring rules are loss functions which assign numerical values to pairs of predictive distributions (given in the form of a PDF, CDF or a discrete sample such as an ensemble forecast) and corresponding validating observations. Our EMOS predictive distributions are based on the continuous ranked probability score \citep[CRPS;][Section 9.5.1]{w19}, which is one of the most popular proper scoring rules in environmental sciences assessing simultaneously both calibration and sharpness of the probabilistic forecast. The former refers to statistical consistency between the forecasts and the corresponding observations, whereas the latter refers to the concentration of the predictive distribution. The CRPS corresponding to a predictive CDF \ $F$ \ and observation \ $x\in {\mathbb R}$ \ is defined as 
\begin{equation}
    \label{eq:CRPSdef}
\crps(F,x) := \int_{-\infty}^{\infty}\Big[F(y)-{\mathbb I}_{\{y\geq x\}}\Big]^2{\mathrm d}y ={\mathsf E}|X-x|-\frac 12
{\mathsf E}|X-X'|,
\end{equation}
where  \ $X$ \ and \ $X'$ \ are independent random variables with CDF \ $F$ \ and finite first moment. The CRPS is a negatively oriented score, that is smaller values imply better forecast skill, and the right hand side of \eqref{eq:CRPSdef} implies that it can be expressed in the same units as the observation. Note that the CRPS for all distribution families considered in Section \ref{subs3.1} can be expressed in closed form \citep[for the corresponding formulae see e.g.][]{jkl19} allowing a computationally efficient optimization procedure. Finally, for point forecasts the CRPS reduces to the absolute error, that is the  mean CRPS over the training data is replaced by the mean absolute error (MAE). In this case the mean squared error (MSE) or the root mean squared error (RMSE) are also popular loss functions.

\subsection{Machine learning-based models}
\label{subs3.3}
Our first approach extends the machine learning-based TN model of \citet{bb21} to the censored Gaussian predictive distribution of Section \ref{subs3.1.2} for calibrating solar irradiance ensemble forecasts, and we also consider a wind speed model with an LN predictive law. Note that initial tests with machine learning-based estimation of the parameters of the censored logistic predictive distribution for calibrating solar irradiance forecasts were also performed. However, probably due to the more complex form of the loss function, the model training was far less stable than in the case of the censored Gaussian law.

\subsubsection{Simple MLP model}
\label{subs3.3.1}
To estimate the unknown parameters of the predictive distributions, an MLP is trained. It is a feedforward neural network, which in our case consists of an input layer, followed by a single hidden layer, whereas the last layer is the output layer consisting of as many neurons as many parameters are to be estimated. In contrast to \citet{rl18}, where besides the mean and standard deviation of the ensemble forecasts of the investigated 2m temperature, station specific information and the ensemble mean and standard deviation of several other weather quantities are used as input features, or to \citet{gzshf22} augmenting ensemble statistics for a given location with data of the neighbouring ones and with station specific information as well, our approach is based merely on some functions of the raw ensemble.

The weights of the network are determined by minimizing the mean CRPS over the training data. Instead of a very long static training set (e.g. \citet{gzshf21} uses daily forecast-observation pairs of 28 years), we consider the same  $\ell_{tp}$-day rolling training periods as in EMOS modelling. However, due to the much larger number of unknown weights of the MLP, the training of the network requires more training data than the EMOS approaches of Section \ref{subs3.1}. To avoid the increase of the training period length, the different forecast periods are pooled and just two networks are trained: one for the 0--24h forecasts, and another one for the 24--48h forecasts. For the implementation details of this initial network, referred to as {\em MLP-S\/}, see Sections \ref{subs4.1.1} and \ref{subs4.2.1}.

\subsubsection{Extended model}
\label{subs3.3.2}
Predictive performance of the individual ensemble forecasts can be very different, sometimes independently of the location and forecast horizon, which highly affects the success of distribution fitting. Typically, a less accurate forecast ensemble results in a wider central prediction interval even in the case of a minor change in the corresponding CRPS value. To correct this deficiency, before training the network for the estimation of the distributional parameters, in our second approach we try to improve first the forecast accuracy. For this aim we introduce two different auxiliary neural networks based on the same input features as the one described in Section \ref{subs3.3.1}, both providing additional corrected point forecasts for each location, time point and forecast horizon. 

One of the newly introduced networks is again an MLP having quite a similar structure as the first network, optionally with more hidden layers. However,  the output layer contains just a single neuron providing a point forecast, and the loss function is the MSE or the MAE. This network, referred to as {\em MLPaux\/}, is trained using forecast-observation pairs of the  $\ell_{tp}$-day rolling training period and used to create in-sample corrected training forecasts and out-of-sample corrected forecast for the corresponding validation time point.

The construction of the other network is different, as we treat forecasts and observations corresponding to subsequent time points as sequences.
For a given location consider separately the  \ $\ell _{tp}L$ \ long time series of the input features and the corresponding observations from the \ $\ell _{tp}$-day training period having \ $L$ \ observations per day. Then each of these sequences is split into shorter, overlapping slices of length \ $\ell_{tw}$, \ considered as time windows. In the case of the features, the first slice is a sequence of vectors consisting of the feature vectors corresponding to the first \ $\ell_{tw}$ \ time steps, while the second slice is obtained by shifting the time window by \ $w$, \ where \ $w\leq \ell_{tw}$, \ and \ $w$ \ and \ $\ell_{tw}$ \ are hyperparameters of the network. For each sequence of feature vectors the corresponding sequence of observations is created in a similar way.

For example, consider a three-dimensional input feature vector \ $\big(x_1^{(t)},x_2^{(t)},x_3^{(t)}\big)^{\top}$, \ where index \ $t$ \ indicates the time of validity of the corresponding forecasts. Assume that the length of slices is  \ $\ell_{tw}=12$, \ and the shift is \ $w=4$. \ Then the first two sequences of features are
\begin{equation*}
\begin{pmatrix}
 x_1^{(1)}\\x_2^{(1)}\\x_3^{(1)}\end{pmatrix},
 \begin{pmatrix}x_1^{(2)}\\x_2^{(2)}\\x_3^{(2)}\end{pmatrix}, \ldots,
\begin{pmatrix}x_1^{(12)}\\x_2^{(12)}\\x_3^{(12)}\end{pmatrix}  \qquad \text{and} \qquad \begin{pmatrix}
 x_1^{(5)}\\x_2^{(5)}\\x_3^{(5)}\end{pmatrix},
 \begin{pmatrix}x_1^{(6)}\\x_2^{(6)}\\x_3^{(6)}\end{pmatrix}, \ldots,
\begin{pmatrix}x_1^{(16)}\\x_2^{(16)}\\x_3^{(16)}\end{pmatrix}.
\end{equation*}
These sequences of length \ $\ell_{tw}$ \ are the inputs of the network, while the target vectors are the same slices of the observations.

The first layer of the network is a so-called 1D convolution layer \citep[see e.g.][]{kaa21} with kernel size \ $\kappa$, \ which means that a filter 
window of width \ $\kappa$ \ is sliding along the given input sequence (the height of the window equals the number of input features), and computes a weighted sum of the underlying values. In this way all \ $\kappa$ \ subsequent feature vectors share the same weight matrix, helping to recognize the temporal behaviour of the process. The number of the filter windows is a hyperparameter of the network. The 1D convolutional layer is followed by a pooling layer, where either average or maximum pooling is performed. This first pooling layer might optionally be followed by further 1D convolution and pooling layers. The next layer is a flatten layer, which simply transforms the data to a vector, the last layer before the output layer is a fully connected one, and the number of neurons in the output layer is equal to the length \ $\ell_{tw}$ \ of the target vectors (and to the length of the input feature sequences as well). 

To summarize, this network, referred to as {\em C1Daux\/}, provides a sequence-to-sequence estimate; from a sequence of feature vectors of length \ $\ell_{tw}$ \ it computes a sequence of forecasts of the same length. As loss function the MSE or the MAE are applied, where the computed forecasts and the verifying observations (the target vectors) are compared.

After training this second network with data from the  $\ell_{tp}$-day training period, we create an in-sample corrected training prediction time series in the following way. We consider the same  \ $\ell _{tp}L$ \ long vector time series of the input features as before and divide it into slices of length \ $\ell_{tw}$ \ again; however, now the slices are not overlapping. The same procedure is repeated with the vector time series corresponding to the \ $L$ \ time points of the validation day, resulting in a corrected out-of-sample forecast.

Finally, to estimate the parameters of the predictive distributions we consider an MLP neural network similar to the one described in Section \ref{subs3.3.1}; however, we extend the set of the input features with the corrected point forecasts obtained using the two auxiliary networks MLPaux and C1Daux. In the following sections, this extended MLP will be referred to as {\em MLPex\/}.

\subsection{Forecast evaluation}
\label{subs3.4}

The general aim of probabilistic forecasting, as formulated by \citet{gbr07}, is to access the maximal sharpness of the predictive distribution subject to calibration.

One of the simplest tools for assessing calibration of ensemble forecasts is the verification rank histogram displaying the ranks of the verifying observations with respect to the corresponding ensemble forecasts \citep[][Section 9.7.1]{w19}. For a properly calibrated $K$-member ensemble each of the \ $K+1$ \ ranks is equally likely resulting in a completely flat uniform histogram. The continuous counterpart of the verification rank histogram for probabilistic forecasts specified by predictive distributions is the probability integral transform (PIT) histogram \citep[][Section 9.5.4]{w19}. PIT is defined as the value of the predictive CDF evaluated at the verifying observation, and for a calibrated predictive distribution PIT values follow a standard uniform law.

However, proper scoring rules allow us to address calibration and sharpness simultaneously. In the case studies of Section \ref{sec4} the predictive performance of the competing probabilistic forecasts with a given forecast horizon is compared with the help of the mean CRPS over all forecast cases in the verification period. Further, the improvement in terms of the mean CRPS of a probabilistic forecast \ $F$ \ with respect to a reference forecast \ $F_{ref}$ \ can be quantified using the continuous ranked probability skill score \citep[CRPSS; see e.g.][]{gr07}.  Denoting by \ $\overline\crps_F$ \ and \ $\overline\crps_{F_{\text{ref}}}$ \ the mean CRPS corresponding to forecasts \ $F$ \ and \ $F_{\text{ref}}$, \ respectively, the CRPSS is defined as
\begin{equation*}
  \crpss := 1 - \frac{\overline\crps_F}{\overline\crps_{F_{\text{ref}}}}.
\end{equation*}
Note that skill scores are positively oriented, that is larger CRPSS means better predictive performance compared to the reference forecast.

Calibration and sharpness of a probabilistic forecast can also be investigated with the help of the coverage and the average width of the \ $(1-\alpha)100\,\%, \ \alpha \in ]0,1[$, \ central prediction intervals, respectively. As coverage we consider the proportion of validating observations located between the lower and upper \ $\alpha/2$ \ quantiles of the predictive distribution, which for a calibrated forecast should be approximately \ $(1-\alpha)100\,\%$. \ In order to provide a fair comparison with a $K$-member ensemble forecast, level \ $\alpha$ \ should be chosen to match the nominal coverage of \ $(K-1)/(K+1)100\,\%$  \ of the ensemble range ($83.33\,\%$ for the $11$-member AROME-EPS).

Finally, point forecasts such as median and mean of the raw ensemble and of
the calibrated predictive distributions are evaluated with the use of MAEs and RMSEs, and one should remark that the MAE optimal for the median, whereas the RMSE is optimal for the mean \citep{gneiting11}.

\section{Case studies}
\label{sec4}

The efficiency of the simple and extended MLP approaches proposed in Sections \ref{subs3.3.1} and \ref{subs3.3.2}, respectively, is tested in two different case studies based on short term AROME-EPS ensemble forecasts of 100m wind speed and GHI introduced in Section \ref{sec2}. In both cases forecast-observation pairs of a whole year from 1 July 2020 to 30 June 2021 are used for model verification, where the forecast skill of the machine learning-based methods is compared to that of the matching EMOS models and the raw ensemble predictions.  As mentioned, in the case of EMOS models each forecast horizon is treated separately, whereas for the MLP-S and MLPEx approaches one network is trained for the 0--24h forecasts and another one for the 24--48h predictions. For wind speed this results in 192 sets of EMOS model parameters corresponding to a given forecast initialization time, whereas for solar irradiance one has 96 different sets.

\subsection{Wind speed}
\label{subs4.1}

Calibration of 100m wind speed ensemble forecasts is performed locally using 51-day rolling-training periods, which training period length is derived from a detailed preliminary data analysis \citep[for details see][]{bb21}.

\subsubsection{Implementation details of the neural networks}
  \label{subs4.1.1}

The input features of the MLP-S network are the control member \ $f_{CTRL}$, \ the mean of the exchangeable members \ $\overline f_{ENS}$, \  and  the ensemble standard deviation \ $S$. \ The same three features are used to train the C1Daux network, while the MLPaux is based on the ensemble mean \ $\overline f$ \ and standard deviation \ $S$. \ The network MLPex has five input features; namely, the same three as MLP-S, extended with the point forecasts provided by MLPaux and C1Daux. 

The main hyperparameters of the different networks are the following. MLP-S and MLPex have the same architecture: one hidden layer with 28 neurons. Here the batch size is 1024, the initial learning rate is 0.01, which is halved in the epochs 8, 28, 48, 68, and the Adam optimizer is used. In the output layer there are 2 neurons providing the estimation of the values \ ${\mathrm e}^\mu$ \ and \ ${\mathrm e}^\sigma$, \ resulting in the location \ $\mu$ \ and the scale \ $\sigma$ \ of the predictive distribution, respectively. Finally, in the input and hidden layers the {\em elu\/}, whereas in the output layer the linear activation function is applied.  MLPaux has two hidden layers with 5 and 15 neurons, respectively, the batch size is 1024, the initial learning rate is 0.01, and it is multipled by 0.97 in the epochs  \ $3,4,\ldots ,59$. \ The network is trained by minimizing the mean absolute error using the Adam optimizer. The C1Daux has a 1D convolutional layer with 24 filters, the kernel size \ $\kappa$ \ is 3. The next layer is a max pooling layer with a pool size of 2, which is followed by a flatten layer and a dense layer with 25 neurons. The batch size is 512, while the learning rate, the loss function and the optimizer are the same as for the MPLaux. Considering the input sequences, \ $\ell _{tw}=16$ \ and  \ $w=4$ \ (see Section \ref{subs3.3.2}). The auxiliary networks use the linear activation function in the output layer and the {\em relu\/} function in the other layers.

For all networks the training data is randomly divided into two parts, $80\,\%$ of the data is used for optimization of the network's parameters, the remaining $20\,\%$ for the stopping criterion. After each epoch the loss function is evaluated on this data set, and if this value is increasing in 10 subsequent epochs, then the training will terminate.

\subsubsection{Post-processing of 100m wind speed ensemble forecasts}
\label{subs4.1.2}

\begin{figure}[t]
  \centering
\epsfig{file=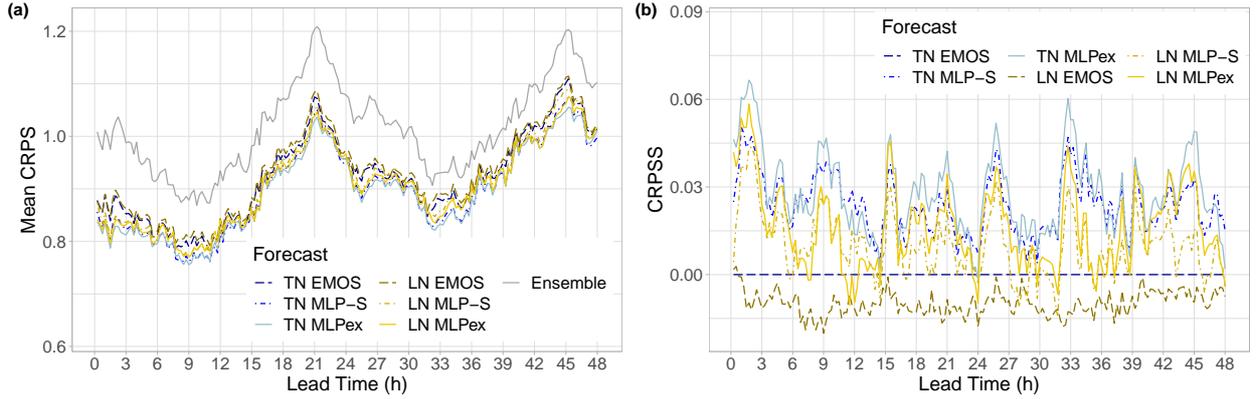, width=\textwidth}
\caption{Mean CRPS of post-processed and raw 100m wind speed ensemble forecasts (a) and CRPSS of post-processed forecasts with respect to the TN EMOS model (b) as functions of the lead time.}
\label{fig:crps_leadW}
\end{figure}

\begin{table}[t]
  \begin{center}
\begin{tabular}{c|c|c|c|c|c}
  TN EMOS&TN MLP-S&TN MLPex&LN EMOS&LN MLP-S&LN MLPex\\ \hline
  90.44\,\%&88.30\,\%&87.91\,\%&91.33\,\%&89.55\,\%&88.94\,\%
\end{tabular}
\end{center}
\caption{Overall mean CRPS of post-processed 100m wind speed forecasts as proportion of the mean CRPS of the AROME-EPS}
\label{tab1}
\end{table}
  
According to Figure \ref{fig:crps_leadW}a, in terms of the mean CRPS all considered post-processing models outperform the raw 100m wind speed ensemble forecasts for all lead times by a wide margin. A better insight into the differences between the various calibration approaches can be obtained from Figure \ref{fig:crps_leadW}b showing the CRPSS values with respect to the TN EMOS model. In general, machine learning-based methods outperform the corresponding EMOS models for all lead times and in most cases MLPex results in higher skill score than the matching MLP-S. According to Table \ref{tab1},
providing the overall mean CRPS of post-processed forecasts as proportion of the mean CRPS of the AROME-EPS, TN MLPex results in the lowest mean score value closely followed by the TN MLP-S, LN MLPex and LN MLP-S.

\begin{figure}[t]
  \centering
\epsfig{file=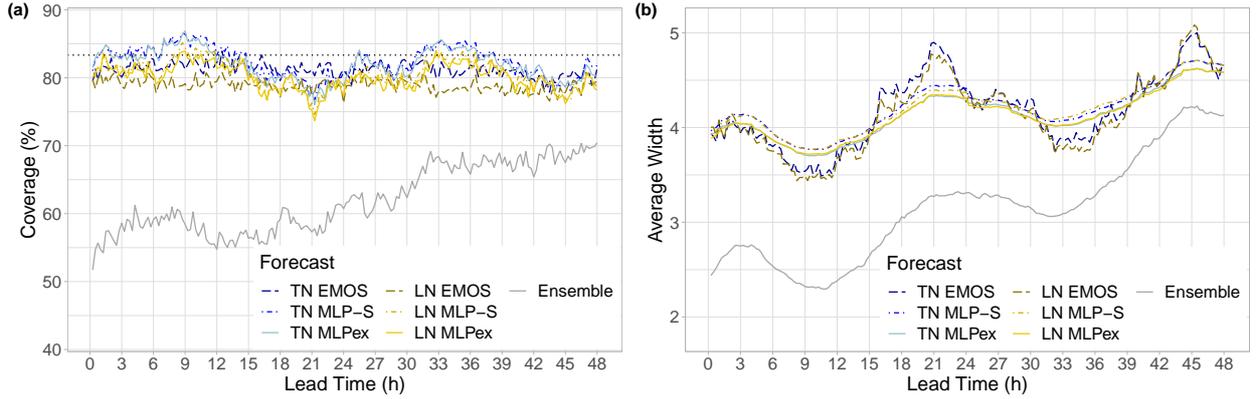, width=\textwidth}
\caption{Coverage (a) and average width (b) of the nominal 83.33\,\% central prediction intervals of post-processed and raw 100m wind speed ensemble  forecasts as functions of the lead time.}
\label{fig:cov_leadW}
\end{figure}

\begin{table}[t]
  \begin{center}
\begin{tabular}{c|c|c|c|c|c|c}
  TN EMOS&TN MLP-S&TN MLPex&LN EMOS&LN MLP-S&LN MLPex&Ensemble\\ \hline
  2.18\,\%&1.91\,\%&2.09\,\%&4.44\,\%&2.91\,\%&3.29\,\%&21.30\,\%
\end{tabular}
\end{center}
\caption{Mean absolute deviation in coverage from the nominal 83.33\,\% level over all lead times.}
\label{tab2}
\end{table}

In Figure \ref{fig:cov_leadW} the coverage and the average width of nominal 83.33\,\% central prediction intervals are displayed as functions of the forecast horizon. All post-processed forecasts result in a coverage rather close to the nominal level, whereas the coverage of the raw ensemble is between 51.72\,\% and 70.43\,\% and shows an increasing trend as the lead time increases (see Figure \ref{fig:cov_leadW}a). The mean absolute deviations from the nominal coverage of 83.33\,\% are summarized in Table \ref{tab2}, which provides a possible ranking of the different forecasts. Obviously, the improved coverage of post-processed forecasts is a result of wider central prediction intervals, as depicted in Figure \ref{fig:cov_leadW}b. All machine learning approaches provide smooth curves of average width, which is a result of using just two different sets of trained networks, whereas in EMOS modelling each lead time is considered separately. From the MLP models the extended ones result in the sharpest predictive distributions outperforming the EMOS methods for lead times 15h to 31h and 41h to 46h.

\begin{figure}[t!]
  \centering
\epsfig{file=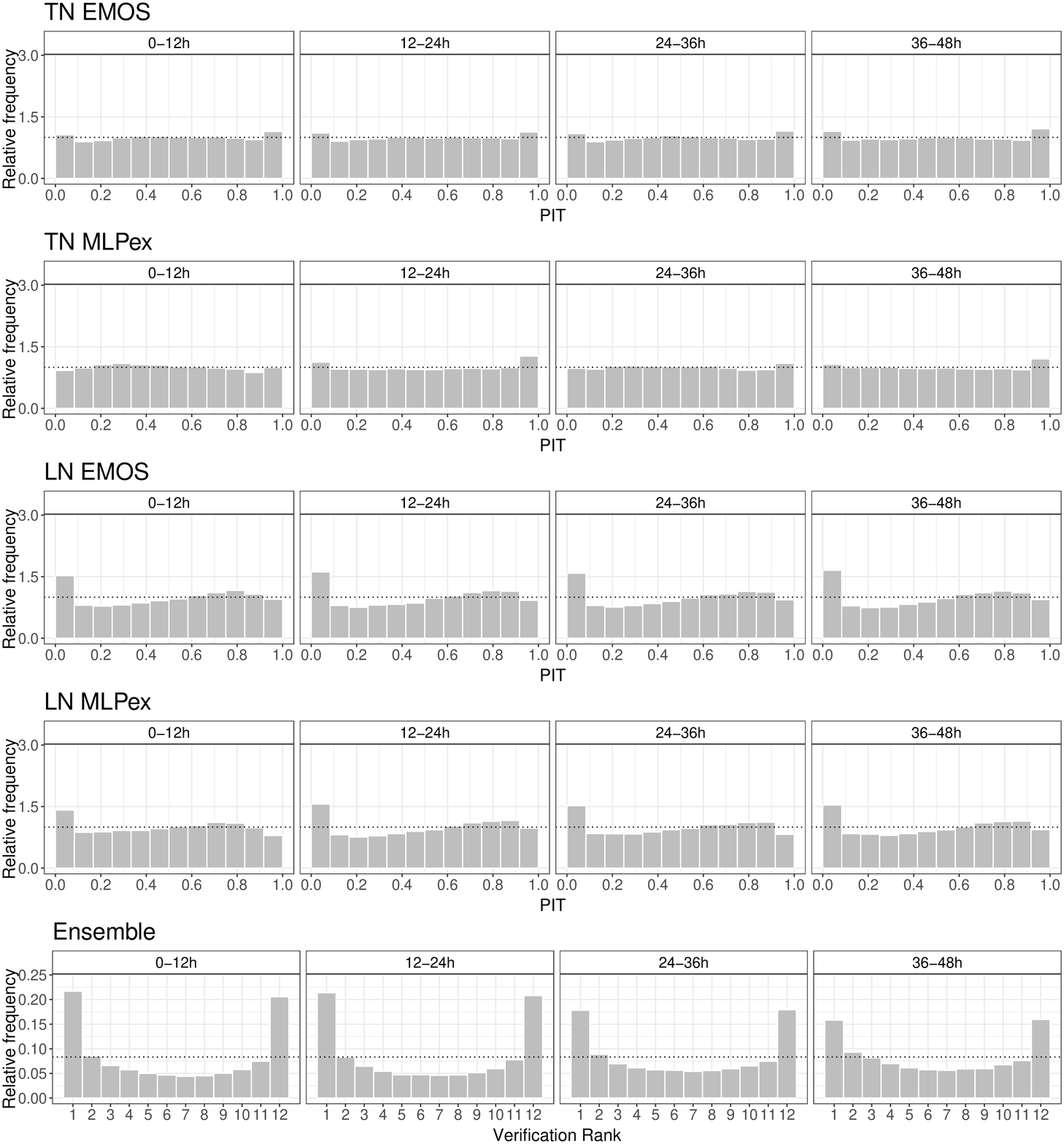, width=\textwidth}
\caption{PIT histograms of post-processed and verification rank histograms of raw 100m wind speed ensemble forecasts for the lead times 0--12h, 12--24h, 24--36h and 36--48h.}
\label{fig:pitW}
\end{figure}

The better calibration of post-processed forecasts can also be observed on Figure \ref{fig:pitW} showing their PIT histograms together with the verification rank histograms of the raw ensemble corresponding to four different lead time intervals. Note that as there are no visible differences between the PIT histograms of the MLPex predictive distributions and the corresponding MLP-S approaches, the latter ones are not shown. The verification rank histograms of the AROME-EPS are symmetric and rather U-shaped. The former indicates the lack of bias in the ensemble forecasts, whereas the latter is a sign of a strongly underdispersive character; however, the dispersion improves with the forecast lead time. The underdispersion is nicely corrected by post-processing, especially when a TN predictive distribution is utilized.
For models with an LN predictive distribution the moment-based $\alpha^0_{1234}$ test \citep{k15} rejects uniformity at a 5\,\% level of significance for all available lead times, whereas for the TN EMOS, TN MLP-S and TN MLPex approaches the acceptance rate is 36.5\,\%, 47.4\,\% and 40.6\,\%, respectively.

\begin{figure}[t]
  \centering
\epsfig{file=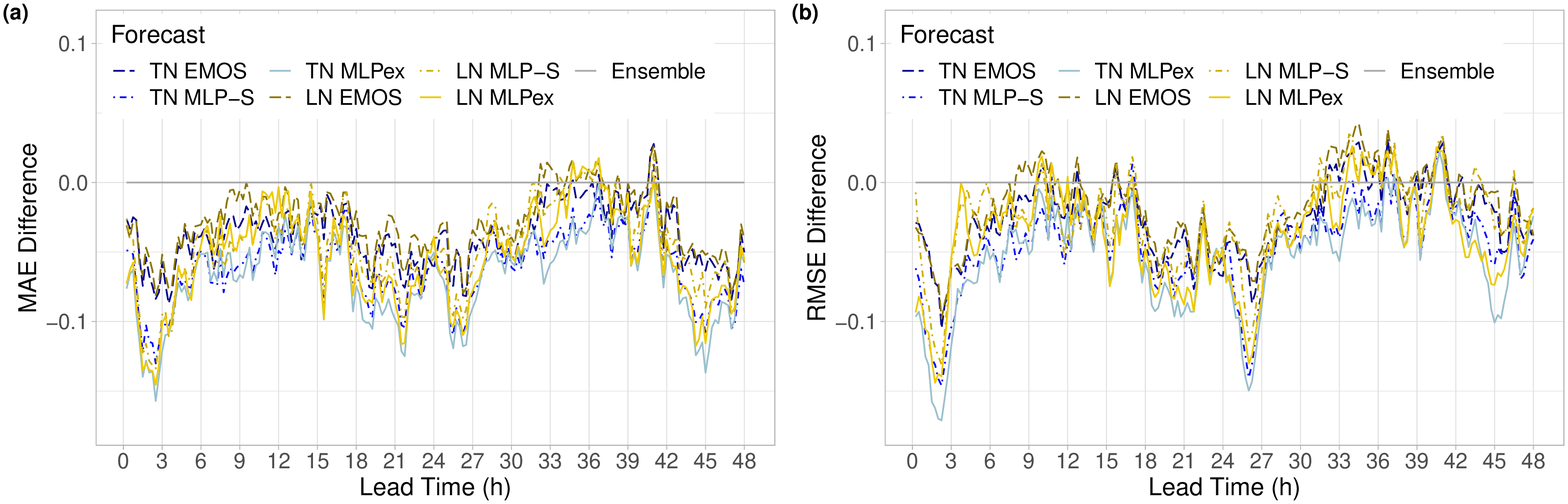, width=\textwidth}
\caption{Difference in MAE of the median forecasts (a) and in RMSE of the mean forecasts (b) from the raw ensemble as functions of the lead time.}
\label{fig:maed_leadW}
\end{figure}

Finally, according to  Figure \ref{fig:maed_leadW}, where the difference in MAE of the median forecasts and in RMSE of the mean forecasts from the raw AROME-EPS are depicted as functions of the forecast horizon, post-processing does not really improve the accuracy of point predictions. This is in line with the findings of \citet{bb21} and might be explained by the lack of bias on the ensemble forecasts, meaning no further bias correction is required.

On the basis of the above results one can conclude that for both investigated predictive distributions the machine learning-based approaches outperform their EMOS counterparts and the extended MLP exhibits slightly better forecast skill than the simple MLP. For the data set at hand the best performing post-processing methods are the novel TN MLPex and the TN MLP-S resulting in the lowest CRPS, the best coverage and the most uniformly distributed PIT values.

\subsection{Solar irradiance}
\label{subs4.2}

In contrast to wind speed, AROME-EPS forecasts of GHI are calibrated regionally allowing extrapolation of the investigated models to sites excluded from the training. According to the data analysis performed by \citet{sealb21}, a 31-day rolling training period is applied.

  \subsubsection{Implementation details of the neural networks}
  \label{subs4.2.1}
The MLP-S network applied for calibrating solar irradiance ensemble forecasts is trained by five input features, which are the control member \ $f_{CTRL}$, \ the mean of the exchangeable members \ $\overline f_{ENS}$, \  the ensemble standard deviation \ $S$, \ an integer between $0$ and $47$ corresponding to the forecast horizon and  the proportion \ $p_0$ \ of zero forecasts in the ensemble.  The  MLPaux network is based on the same features as the MLP-S but \ $p_0$, \ whereas the C1Daux network uses only the mean and the standard deviation of the 11-member ensemble.  Finally, the net MLPex works with seven input features; besides the five considered also by MLP-S, it uses the corrected point forecasts provided by the two auxiliary networks.

The structures of the networks are the following. Both MLP-S and MLPex have only one hidden layer with 35 neurons and exponential activation function. In the output layer there are two neurons providing the estimation of the values \ $\mu ^3$ \ and \ $e^\sigma$, \ and here the activation function is the linear one. MLPaux has one hidden layer with 32 neurons and with a relu activation function, followed by a normalization layer, and by the output layer with a linear activation function. The C1Daux has a 1D convolutional layer with 35 filters, the kernel size \ $\kappa$ \ equals 5. The next layer is an average pooling layer with a pool size of 2, followed by a second 1D convolutional layer with 16 filters and kernel size \ $\kappa =2$, a flatten layer and a dense layer with 30 neurons. Here the input sequences are of length \ $\ell _{tw}=12$ \ and the shift is \ $w=1$. 

\subsubsection{Post-processing of GHI ensemble forecasts}
\label{subs4.2.2}

\begin{figure}[t]
  \centering
\epsfig{file=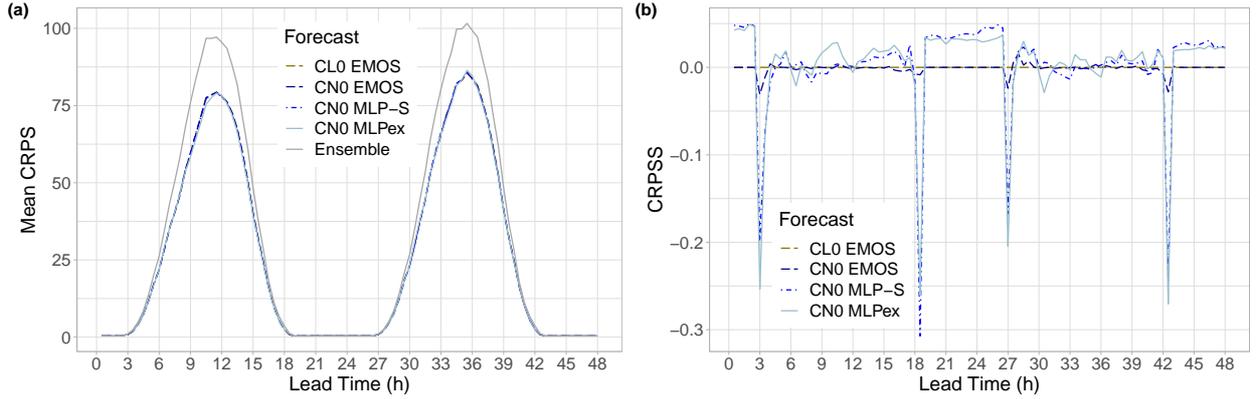, width=\textwidth}
\caption{Mean CRPS of post-processed and raw GHI ensemble forecasts (a) and CRPSS of post-processed forecasts with respect to the CL0 EMOS model (b) as functions of the lead time.}
\label{fig:crps_leadR}
\end{figure}

Figure \ref{fig:crps_leadR} shows the mean CRPS of post-processed and raw GHI ensemble forecasts and the CRPSS with respect to the CL0 EMOS model suggested by \citet{sealb21}. Compared with the raw AROME-EPS ensemble, all calibration methods result in a substantial decrease in the mean CRPS between 3 and 19 UTC when positive GHI is likely to be observed. According to Figure \ref{fig:crps_leadR}b, there is just a minor difference in skill between the CL0 and CN0 EMOS models, while the two MLP-based approaches result in positive CRPSS in more than 75\,\% of the considered lead times and perform particularly well in the ``dark'' hours (0--3 and 19--24 UTC). However, from the point of view of PV energy production, one is interested in the predictive performance of the various forecasts when the observed GHI exceeds some positive threshold. To address this problem, Table \ref{tab3} summarizes the mean CRPS of post-processed forecasts as proportion of the mean CRPS of the AROME-EPS for forecast cases with observed irradiance not less than 7.5 $W/m^2$, which threshold had been suggested by forecasters of the HMS. For these cases the highest skill corresponds to the CN0 MLPex approach, followed by the CN0 MLP-S, which is still more than 2.8\,\% ahead of the best performing EMOS method.

\begin{table}[t]
  \begin{center}
\begin{tabular}{c|c|c|c}
  CL0 EMOS&CN0 EMOS&CN0 MLP-S&CN0 MLPex\\ \hline
  82.64\,\%&82.67\,\%&79.80\,\%&79.11\,\%
\end{tabular}
\end{center}
\caption{Overall mean CRPS of post-processed GHI forecasts as proportion of the mean CRPS of the AROME-EPS for observed GHI not less than 7.5 $W/m^2$.}
\label{tab3}
\end{table}
 
\begin{figure}[t]
  \centering
\epsfig{file=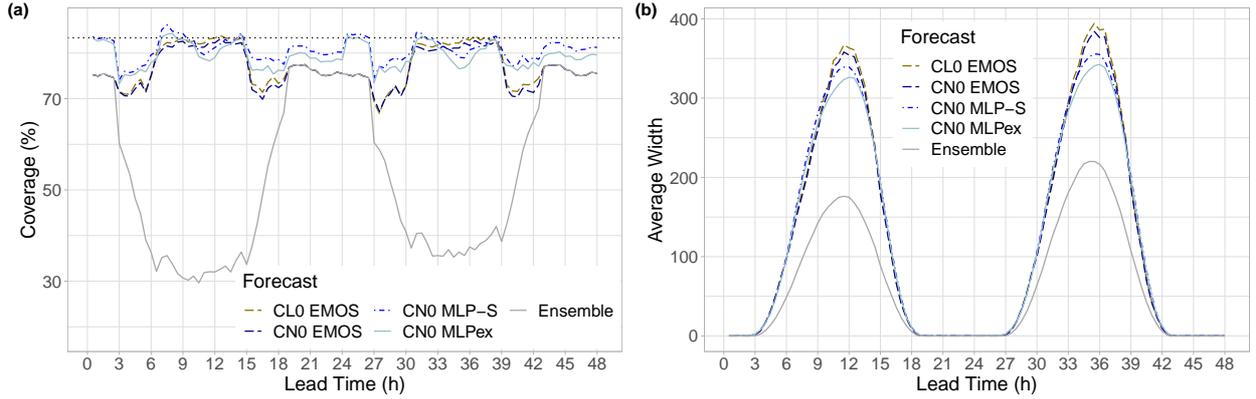, width=\textwidth}
\caption{Coverage (a) and average width (b) of the nominal 83.33\,\% central prediction intervals of post-processed and raw GHI ensemble forecasts as functions of the lead time.}
\label{fig:cov_leadR}
\end{figure}

The improved calibration of post-processed forecasts can be observed in Figure \ref{fig:cov_leadR}a as well, showing the coverage of the nominal 83.33\,\% central prediction intervals. At the hours of peak irradiance (6--15 UTC) all calibrated forecasts result in a coverage very close to the nominal value, whereas the coverage of the raw AROME-EPS forecasts hardly exceeds 40\,\%. In general, the machine learning-based approaches outperform the EMOS models, which conclusion is also supported by Table \ref{tab4} providing the mean absolute deviation in coverage from the nominal 83.33\,\%. Again, as depicted in Figure \ref{fig:cov_leadR}b, the cost of the higher coverage of calibrated forecasts should be paid in the deterioration of the sharpness. From the competing post-processing methods, between 10 and 14 UTC the CN0 MLPex results in far the narrowest central prediction intervals, followed by the CN0 MLP-S, whereas outside these hours there is no much difference in sharpness between the various approaches.

\begin{table}[t]
  \begin{center}
\begin{tabular}{c|c|c|c|c}
  CL0 EMOS&CN0 EMOS&CN0 MLP-S&CN0 MLPex&Ensemble\\ \hline
  6.11\,\%&6.64\,\%&2.88\,\%&4.03\,\%&29.53\,\%
\end{tabular}
\end{center}
\caption{Mean absolute deviation in coverage from the nominal 83.33\,\% level over all lead times.}
\label{tab4}
\end{table}

\begin{figure}[t!]
  \centering
\epsfig{file=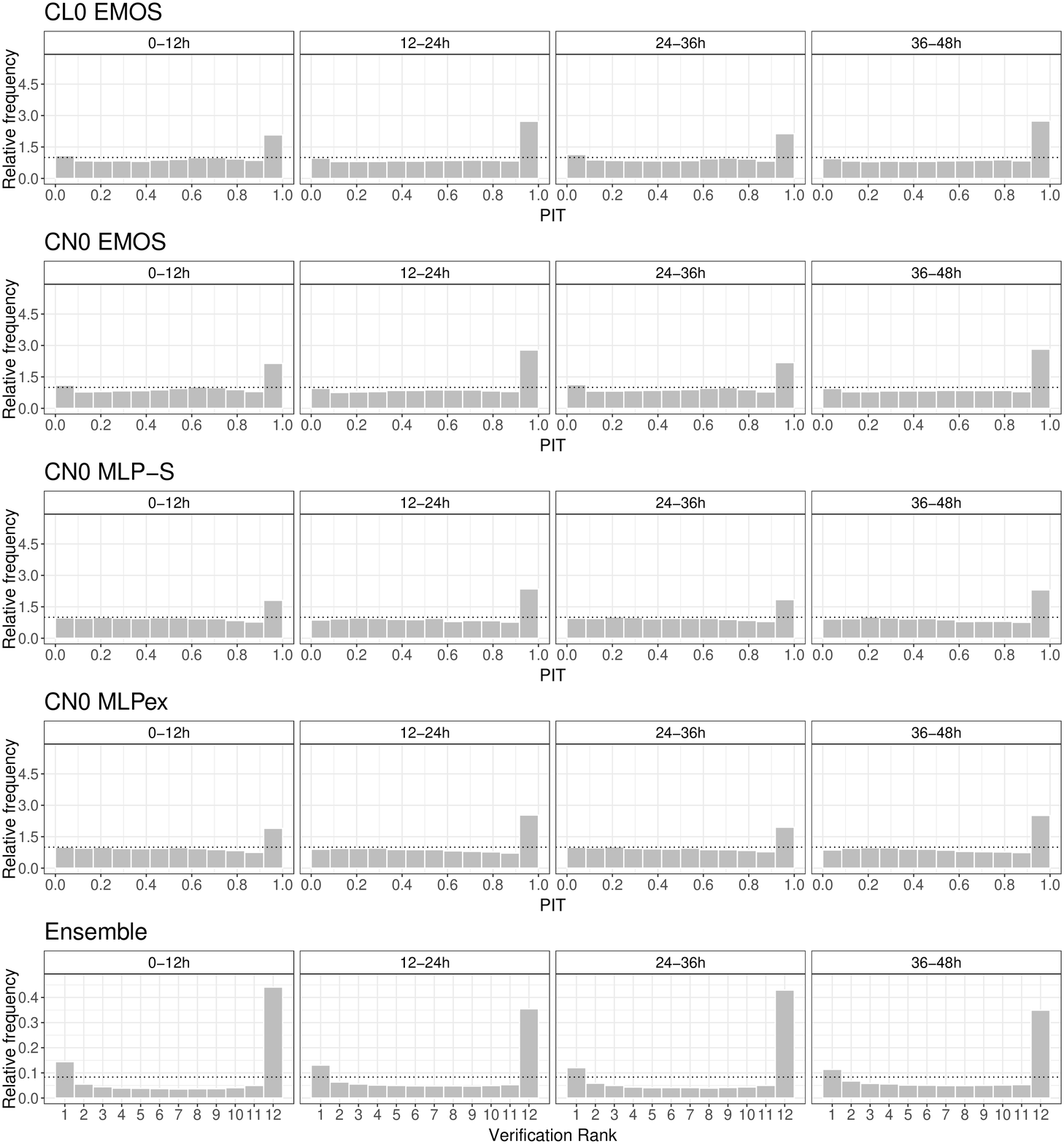, width=\textwidth}
\caption{PIT histograms of post-processed and verification rank histograms of raw GHI ensemble forecasts for the lead times 0--12h, 12--24h, 24--36h and 36--48h.}
\label{fig:pitR}
\end{figure}

Figure \ref{fig:pitR} showing the verification rank histograms of the raw and the PIT histograms of the calibrated GHI ensemble forecasts for lead times 0--12h, \ 12--24h, \ 24--36h and 36--48h also approves the positive effect of post-processing. The raw AROME-EPS forecasts seem to be negatively biased tending to underestimate the observed GHI and indeed, the average biases of the ensemble median and the ensemble mean taken over all lead times and forecast cases with observed GHI not less than 7.5 $W/m^2$ are -16.5 $W/m^2$ and -20.8 $W/m^2$, respectively. Further, the U-shape of the verification rank histograms indicate strong underdispersion, which is in complete accordance with the low coverage values and sharp prediction intervals observed in Figure \ref{fig:cov_leadR}. All post-processing approaches substantially improve the calibration and reduce the bias; however, compared to the case of wind speed (Section \ref{subs4.1.2}), there are much less lead times, where the $\alpha^0_{1234}$ test accepts uniformity at a 5\,\% level of significance. The highest acceptance rate of 4.2\,\% corresponds to the CN0 MLP-S approach (14h, 14.5h, 38h, 38.5h; observations at 14 and 14:30 UTC), followed by the CL0 EMOS (38h, 38.5h) and the CN0 MLPex (38h), whereas for the CN0 EMOS there is no lead time with significantly uniform PIT. 

\begin{figure}[t]
  \centering
\epsfig{file=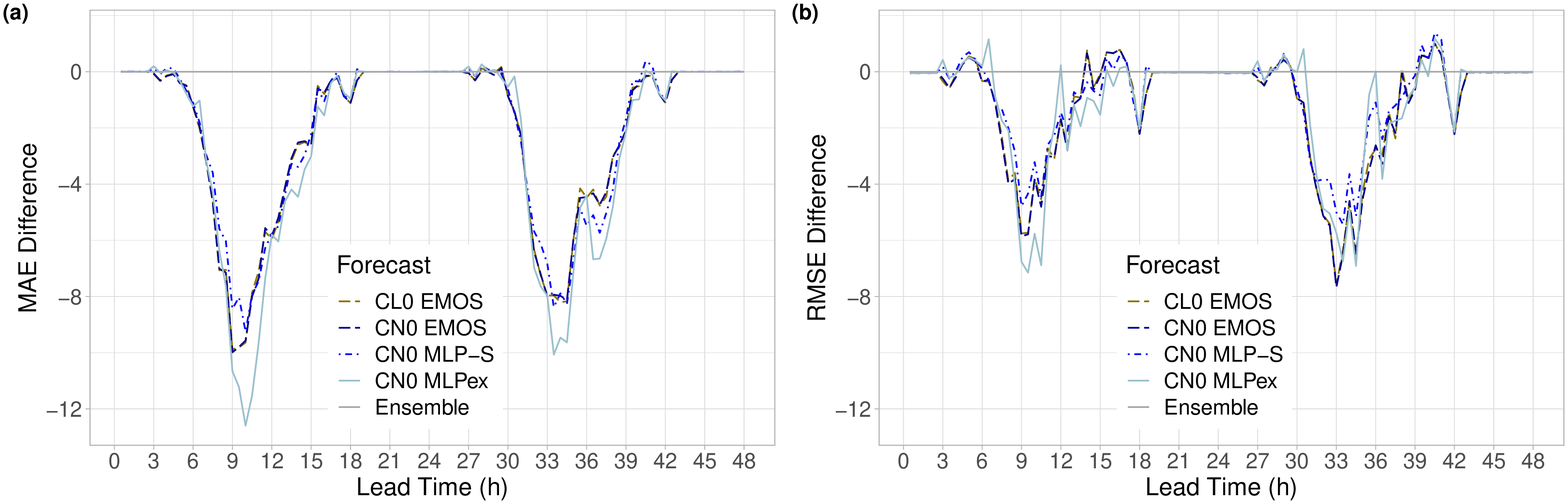, width=\textwidth}
\caption{Difference in MAE of the median forecasts (a) and in RMSE of the mean forecasts (b) from the raw ensemble as functions of the lead time.}
\label{fig:maed_leadR}
\end{figure}

Finally, in Figure \ref{fig:maed_leadR}a the difference in MAE of the median forecasts of the various post-processing methods from the MAE of the raw ensemble are plotted as functions of the forecast horizon. Between 9 and 15 UTC all four investigated approaches result in a substantial improvement in accuracy and the CN0 MLPex method seems to lead to the largest gain in MAE. Fairly similar conclusions can be drawn from  Figure \ref{fig:maed_leadR}b showing the difference in RMSE of the median forecasts and both figures support the presence of a bias in the AROME-EPS which is then corrected by post-processing.

Similar to Section \ref{subs4.1.2}, the results provided above confirm the superiority of the machine learning-based methods over EMOS modelling. The CN0 MLPex approach results in the lowest mean CRPS, the largest improvement in MAE combined with a fair coverage and at the hours around 12 UTC far narrower central prediction intervals then its competitors. 

\subsubsection{Model verification for additional locations}
\label{subs4.2.3}

All approaches to post-processing GHI ensemble forecasts investigated in Section \ref{subs4.2.2} are trained regionally using forecast-observations pairs provided by the HMS for seven locations. Hence, one can use the obtained EMOS models and trained neural networks to calibrate AROME-EPS ensemble forecasts for the solar farms in Monor and Nagyk\H or\"os, data of which are not used in the training process. The raw and post-processed forecasts for these additional locations are then validated against observations 
provided by the solar farm operators, which are, as mentioned in Section \ref{sec2}, in quality far behind the ones provided by the observation network of the HMS. In this way, using the same one year verification period from 1 July 2020 to 30 June 2021, one can investigate the robustness of the various approaches. For simplicity, in this section we investigate the performance of the most skillful EMOS (CL0 EMOS) and machine learning-based (CN0 MLPex) methods only.

\begin{figure}[t]
  \centering
\epsfig{file=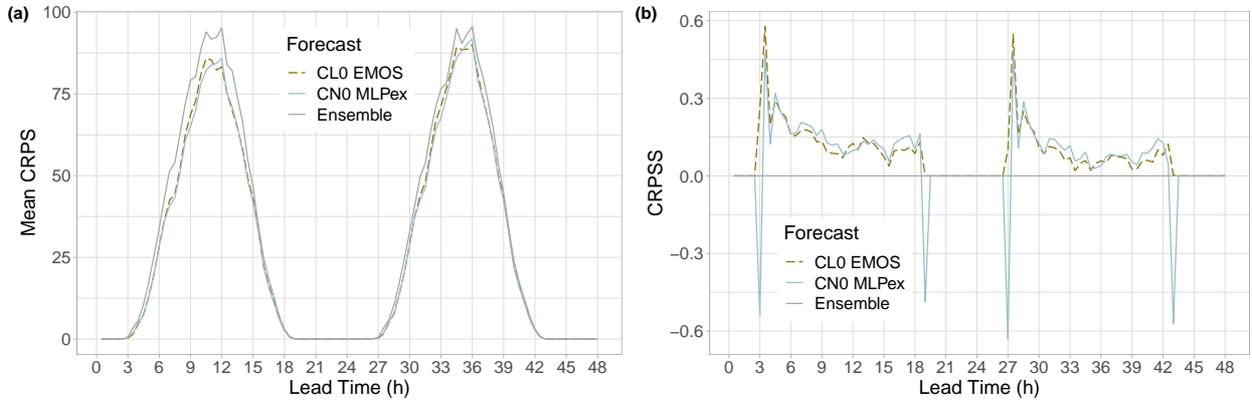, width=\textwidth}
\caption{Mean CRPS of post-processed and raw GHI ensemble forecasts for Monor and Nagyk\H or\"os (a) and CRPSS of post-processed forecasts with respect to the raw ensemble (b) as functions of the lead time.}
\label{fig:crps_lead_PartR}
\end{figure}

Figure \ref{fig:crps_lead_PartR}a again provides the CRPS of post-processed and raw GHI forecasts, whereas in Figure \ref{fig:crps_lead_PartR}b the CRPSS with respect to the raw AROME-EPS is depicted. Compared to Figure \ref{fig:crps_leadR}a, the extrapolated models obviously result in smaller improvements in skill. For forecast cases with observed GHI not less than 7.5 $W/m^2$ the mean CRPS values of the CL0 EMOS and CN0 MLPex approaches are 90.39\,\% and 88.91\,\% of the mean CRPS of the AROME-EPS, that is 7.75\,\% and 9.80\,\% behind the corresponding proportions of Table \ref{tab3}. In general, the machine learning-based CL0 MLPex approach outperforms the CL0 EMOS model, between 6 and 19 UTC it results a higher skill score in more than 71\,\% of the corresponding lead times.

\begin{figure}[t]
  \centering
\epsfig{file=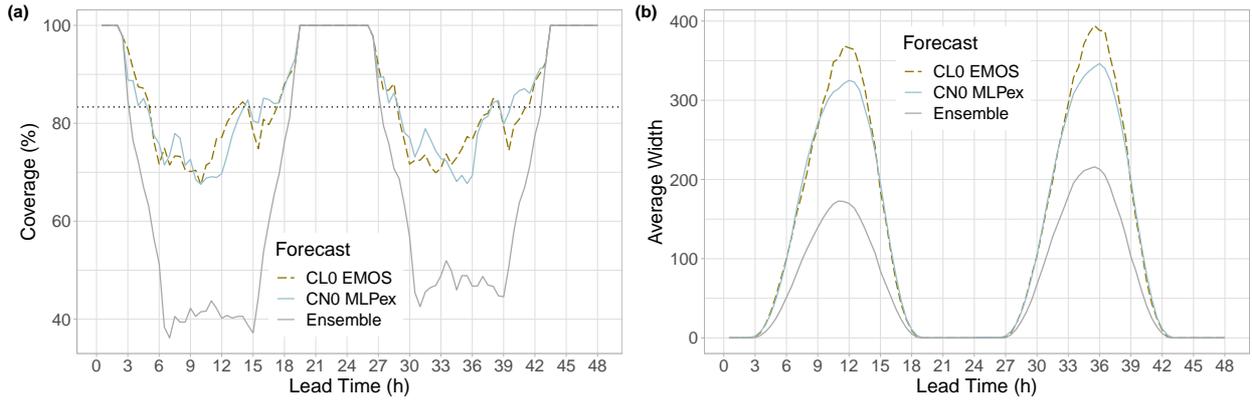, width=\textwidth}
\caption{Coverage (a) and average width (b) of the nominal 83.33\,\% central prediction intervals of post-processed and raw GHI ensemble forecasts for Monor and Nagyk\H or\"os as functions of the lead time.}
\label{fig:cov_lead_PartR}
\end{figure}

Figure \ref{fig:cov_lead_PartR} providing the coverage and average width of the nominal central prediction intervals shows the same general picture as  Figure \ref{fig:cov_leadR}. Compared with the raw ensemble, post-processing substantially improves the calibration at the cost of loss in sharpness, although for the calibrated forecasts the deviation from the nominal 83.33\/\% is higher than in Figure \ref{fig:cov_leadR}a. From the two investigated approaches the novel CL0 MLPex results in slightly better coverage and sharper central prediction intervals.

\begin{figure}[t!]
  \centering
\epsfig{file=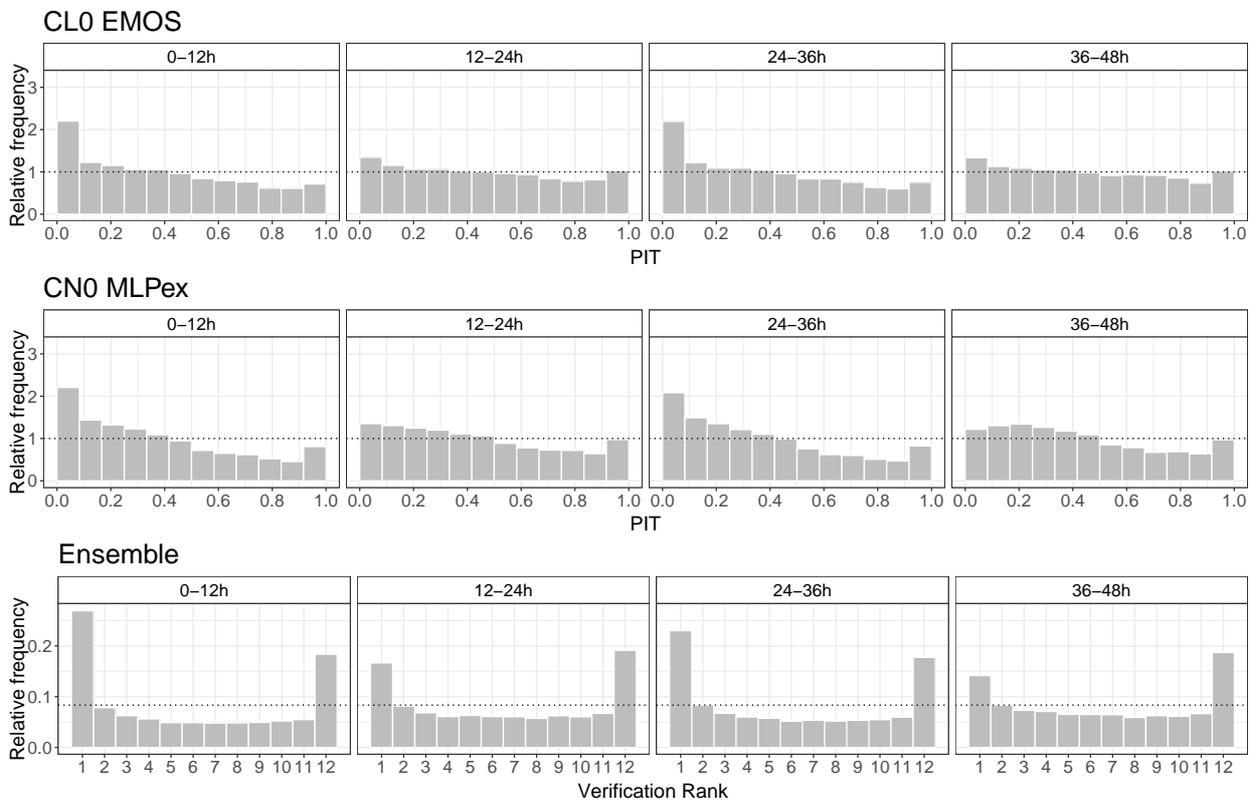, width=\textwidth}
\caption{PIT histograms of post-processed and verification rank histograms of raw GHI ensemble forecasts for Monor and Nagyk\H or\"os for the lead times 0--12h, 12--24h, 24--36h and 36--48h.}
\label{fig:pit_PartR}
\end{figure}

Further, according to the verification rank histograms in the bottom panel of Figure \ref{fig:pit_PartR}, AROME-EPS forecasts for the solar farms at Monor and Nagyk\H or\"os are far less underdispersive than for the seven locations used for model training and exhibit a smaller bias (for observed GHI not less than 7.5 $W/m^2$ the average biases of the ensemble median and mean are 14  $W/m^2$ and 17.9  $W/m^2$, respectively). Both investigated post-processing methods result in more uniform, but slightly biased PIT histograms, especially for lead times 0--12h and 24--36h. In this case, for the CL0 MLPex method the $\alpha^0_{1234}$ test reject uniformity at a 5\,\% level of significance for all available lead times, whereas the CL0 EMOS model has an overall acceptance rate of 32.3\,\%.

\begin{figure}[t]
  \centering
\epsfig{file=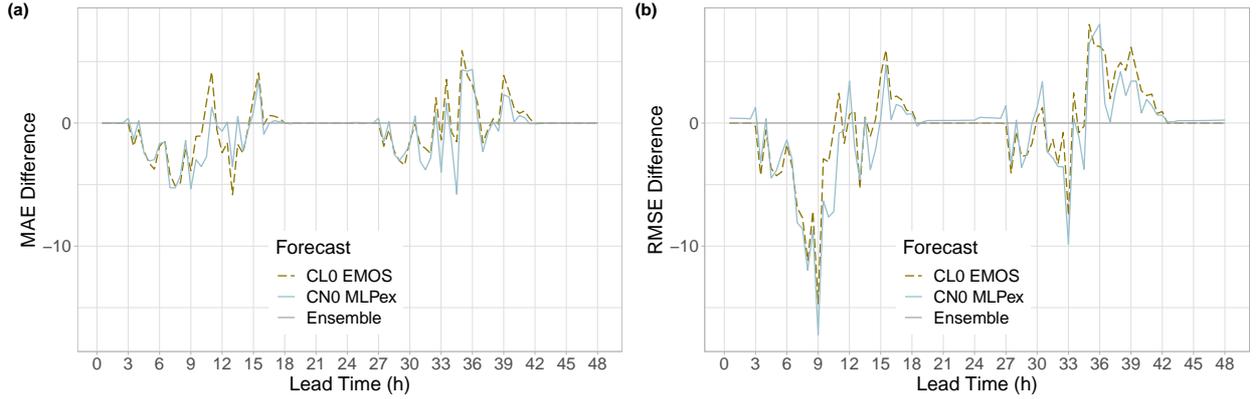, width=\textwidth}
\caption{Difference in MAE of the median forecasts for Monor and Nagyk\H or\"os (a) and in RMSE of the mean forecasts (b) from the raw ensemble as functions of the lead time.}
\label{fig:maed_lead_PartR}
\end{figure}

Finally, as the differences in MAE of the median forecasts and RMSE of the mean forecasts from the raw ensemble plotted in Figure \ref{fig:maed_lead_PartR} indicate, for the two solar farms the improvement in the accuracy of point forecasts is not so consistent as before (see  Figure \ref{fig:maed_leadR}). 

\section{Conclusions}
\label{sec5}

We propose a general two-step machine learning-based approach (MLPex) to calibrating ensemble weather predictions resulting in probabilistic forecasts in the form of full predictive distributions. In the first step, with the help of an MLP neural network and a 1D convolutional neural network we provide improved point forecasts of the investigated weather quantity. These forecasts, augmented with simple statistics of the ensemble predictions, are then serve as input features for an other neural network resulting in the parameters of the predictive distribution. In two case studies, based on 11-member AROME-EPS forecasts of 100m wind speed and global horizontal irradiance of the HMS, the forecast skill of the suggested MLPex method is compared with the predictive performance of the parametric approach using a single MLP to estimate parameters of the predictive distribution (MLP-S), to the state-of-the-art EMOS models and to the raw ensemble forecasts. Only short-range predictions up to 48h lead time are considered with 15-minute and 30-minute temporal resolutions for wind speed and GHI, respectively. Both investigated weather quantities are of great importance in renewable energy production and the considered forecast lead times are of magnitude of the time steps indicated in the scheduling requirements for power plants.

Our case studies confirm that statistical post-processing consistently improves the calibration of probabilistic forecasts resulting in lower mean CRPS values, better coverage of the nominal 83.33\,\% central prediction intervals and PIT histograms far closer to the uniform distribution than the corresponding verification rank histograms of the raw ensemble. The novel two-step MLPex approach exhibits the best overall performance in all situations, followed by the corresponding MLP-S method and the EMOS model. In the case of solar irradiance the robustness of the best performing EMOS and machine learning-based approach is also investigated by applying regionally trained models to calibration of forecasts for two external locations not used in the training process. This extrapolation does not change the ranking of the forecasts; though the gap in skill between the raw and post-processed predictions is smaller than in the case when the same stations are used both for training and verification.

A general advantage of the investigated machine learning-based methods is that they are based on the same type of training data as the corresponding EMOS model. MLPex and MLP-S methods require neither long training periods as the approach of \citet{gzshf21}, nor additional covariates such as forecasts of other weather quantities and/or station specific data, as suggested by \citet{rl18} and \citet{gzshf22}. However, the simple and straightforward opportunity of involving additional input features is not excluded either, providing a possible direction of future research.

Another potential avenue of further studies is to consider multivariate post-processing methods providing temporally consistent forecast trajectories. In this context the MLPex and MLP-S forecasts can serve as initial independent predictions for two-step approaches, where after univariate calibration the temporal dependence is restored with the help of an empirical copula calculated using e.g. the actual ensemble forecasts \citep[ensemble copula coupling;][]{stg13} or historical observations \citep[Schaake shuffle;][]{cghrw04}. For a detailed comparison of the state-of-the-art multivariate methods with the help of simulated and real ensemble data we refer to \citet{lbm20} and \citet{llhb22}, respectively.

\bigskip
\noindent
{\bf Acknowledgments.} \  The work leading to this paper was done in part
during the visit of S\'andor Baran to the Heidelberg Institute for Theoretical Studies in July 2022 as guest researcher. He was also supported by the National Research, Development and Innovation Office under Grant No. NN125679. The authors thank Gabriella Sz\'epsz\'o and Mih\'aly Sz\H ucs from the HMS for providing the AROME-EPS data.

\end{document}